\documentclass[10pt,twocolumn,letterpaper]{article}

\usepackage[pagenumbers]{cvpr}

\usepackage{graphicx}
\usepackage{amsmath}
\usepackage{amssymb}
\usepackage{booktabs}
\usepackage{times}
\usepackage{epsfig}
\usepackage{xcolor}
\usepackage{comment}
\usepackage{multirow}
\usepackage{boldline}
\usepackage{algpseudocode}
\usepackage{algorithm}

\usepackage[pagebackref,breaklinks,colorlinks]{hyperref}

\usepackage[capitalize]{cleveref}
\crefname{section}{Sec.}{Secs.}
\Crefname{section}{Section}{Sections}
\Crefname{table}{Table}{Tables}
\crefname{table}{Tab.}{Tabs.}

\def\confName{CVPR}
\def\confYear{2023}

\begin{document}

\title{MobileViG: Graph-Based Sparse Attention for Mobile Vision Applications}

\author{Mustafa Munir*\\
The University of Texas at Austin\\
{\tt\small mmunir@utexas.edu}
\and
William Avery*\\
The University of Texas at Austin\\
{\tt\small williamaavery@utexas.edu}
\and
Radu Marculescu\\
The University of Texas at Austin\\
{\tt\small radum@utexas.edu} \\
}
\maketitle

\def\thefootnote{*}\footnotetext{Equal contribution}

\begin{abstract}
Traditionally, convolutional neural networks (CNN) and vision transformers (ViT) have dominated computer vision. However, recently proposed vision graph neural networks (ViG) provide a new avenue for exploration. Unfortunately, for mobile applications, ViGs are computationally expensive due to the overhead of representing images as graph structures. In this work, we propose a new graph-based sparse attention mechanism, Sparse Vision Graph Attention (SVGA), that is designed for ViGs running on mobile devices. Additionally, we propose the first hybrid CNN-GNN architecture for vision tasks on mobile devices, MobileViG, which uses SVGA. Extensive experiments show that MobileViG beats existing ViG models and existing mobile CNN and ViT architectures in terms of accuracy and/or speed on image classification, object detection, and instance segmentation tasks. Our fastest model, MobileViG-Ti, achieves 75.7\% top-1 accuracy on ImageNet-1K with 0.78 ms inference latency on iPhone 13 Mini NPU (compiled with CoreML), which is faster than MobileNetV2x1.4 (1.02 ms, 74.7\% top-1) and MobileNetV2x1.0 (0.81 ms, 71.8\% top-1). Our largest model, MobileViG-B obtains 82.6\% top-1 accuracy with only 2.30 ms latency, which is faster and more accurate than the similarly sized EfficientFormer-L3 model (2.77 ms, 82.4\%). Our work proves that well designed hybrid CNN-GNN architectures can be a new avenue of exploration for designing models that are extremely fast and accurate on mobile devices. Our code is publicly available at \url{https://github.com/SLDGroup/MobileViG}.

\end{abstract}

\section{Introduction}

Artificial intelligence (AI) and machine learning (ML) have had explosive growth in the past decade. In computer vision, the key driver behind this growth has been the re-emergence of neural networks, especially convolutional neural networks (CNNs) and more recently vision transformers \cite{liu2022convnet, ViT}. Even though CNNs trained via back-propagation were invented in the 1980s \cite{lecun1989backpropagation, liu2022convnet}, they were used for more small-scale tasks such as character recognition \cite{lecun1998gradient}. The potential of CNNs to re-shape the field of artificial intelligence was not fully realized until AlexNet \cite{Alexnet2012} was introduced in the ImageNet \cite{Imagenet2015} competition. Further advancements to CNN architectures have been made improving their accuracy, efficiency, and speed \cite{MobileNet, MobileNetv2, Resnet, Densenet, VGGNet}. Along with CNN architectures, pure multi-layer perceptron (MLP) architectures and MLP-like architectures have also shown promise as backbones for general-purpose vision tasks \cite{cyclemlp, mlpmixer, resmlp}

Though CNNs and MLPs had become widely used in computer vision, the field of natural language processing used recurrent neural networks (RNNs), specifically long-short term memory (LSTM), networks due to the disparity between the tasks of vision and language \cite{LSTM}. Though LSTMs are still used, they have largely been replaced with transformer architectures in NLP tasks \cite{vaswani2017attention}. With the introduction of Vision Transformer (ViT) \cite{ViT} a network architecture applicable to both language and vision domains was introduced. By splitting an image into a sequence of patch embeddings an image can be transformed into an input usable by transformer modules \cite{ViT}. One of the major advantages of the transformer architecture over CNNs or MLPs is its global receptive field, allowing it to learn from distant object interactions in images.

\begin{figure}[h]
\centering
\includegraphics[scale=0.58]{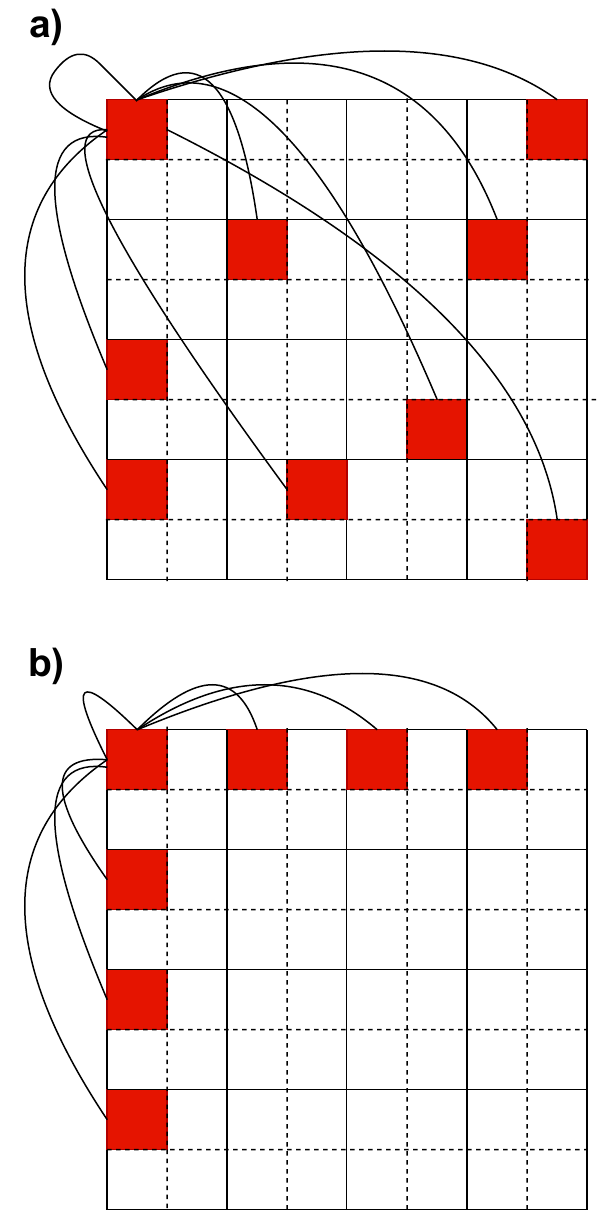}
\caption{a) KNN graph attention for the top left pixel of an 8$\times$8 image as used in Vision GNN. b) SVGA for the top left pixel of an 8$\times$8 image. As shown, SVGA uses a structured graph, that does not change across input images. This structure removes the KNN and reshaping operations required in a) that are not mobile friendly.} 
\label{fig:KNN_ViG}
\end{figure}

Graph neural networks (GNNs) have developed to operate on graph-based structures such as biological networks, social networks, or citation networks \cite{kipf2016semi, hamilton2017inductive, zhou2020graph, wu2020graph}. GNNs have even been proposed for tasks such as node classification \cite{kipf2016semi}, drug discovery \cite{gaudelet2021utilizing}, fraud detection \cite{liu2021pick}, and now computer vision tasks with the recently proposed Vision GNN (ViG) \cite{Vision_GNN}. In short, ViG divides an image into patches and then connects the patches through the K-nearest neighbors (KNN) algorithm \cite{Vision_GNN}, thus providing the ability to process global object interactions similar to ViTs.

\begin{table*}[t]
\caption{MobileViG architecture showing configuration of the stages, output size, downsample layers, and classification head.}
\centering
\begin{tabular}{|c|c|c|c|c|c|}
\hline
Stage                    & Output Size                & MobileViG-Ti & MobileViG-S & MobileViG-M & MobileViG-B \\ \hline \rule{0pt}{4ex}
Stem                     & $\dfrac{H}{4} \times \dfrac{W}{4}$                  & Conv $\times$2             & Conv $\times$2             & Conv $\times$2            & Conv $\times$2           \\[8pt] \hline \rule{0pt}{4ex}
Stage 1                  & $\dfrac{H}{4} \times \dfrac{W}{4}$                 & $\left[ \begin{array}{cc} MBConv \\C=42\end{array}\right]$ $\times$2       & $\left[ \begin{array}{cc} MBConv \\C=42\end{array}\right]$ $\times$3             & $\left[ \begin{array}{cc} MBConv \\C=42\end{array}\right]$ $\times$3            & $\left[ \begin{array}{cc} MBConv \\C=42\end{array}\right]$ $\times$5            \\[8pt] \hline \rule{0pt}{4ex}
$\downarrow$               & $\dfrac{H}{8} \times \dfrac{W}{8}$                  & Conv             & Conv            & Conv            & Conv            \\[8pt] \hline \rule{0pt}{4ex}
Stage 2                  & $\dfrac{H}{8} \times \dfrac{W}{8}$                  & $\left[ \begin{array}{cc} MBConv \\C=84\end{array}\right]$ $\times$2             & $\left[ \begin{array}{cc} MBConv \\C=84\end{array}\right]$ $\times$3            & $\left[ \begin{array}{cc} MBConv\ \\C=84\end{array}\right]$ $\times$3            & $\left[ \begin{array}{cc} MBConv \\C=84\end{array}\right]$ $\times$5            \\[8pt] \hline \rule{0pt}{4ex}
$\downarrow$               & $\dfrac{H}{16} \times \dfrac{W}{16}$                  & Conv             & Conv             & Conv            & Conv             \\[8pt] \hline \rule{0pt}{4ex}
Stage 3                  & $\dfrac{H}{16} \times \dfrac{W}{16}$                  & $\left[ \begin{array}{cc} MBConv \\C=168\end{array}\right]$ $\times$6             & $\left[ \begin{array}{cc} MBConv \\C=176\end{array}\right]$ $\times$9            & $\left[ \begin{array}{cc} MBConv \\C=224\end{array}\right]$ $\times$9             & $\left[ \begin{array}{cc} MBConv \\C=240\end{array}\right]$ $\times$15             \\[8pt] \hline \rule{0pt}{4ex}
$\downarrow$              & $\dfrac{H}{32} \times \dfrac{W}{32}$                  & Conv             & Conv            & Conv             & Conv            \\[8pt] \hline \rule{0pt}{5.3ex}
Stage 4 & $\dfrac{H}{32} \times \dfrac{W}{32}$ & $\left[ \begin{array}{cc} SVGA\\K=2 \\C=256\end{array}\right]$ $\times$2             & $\left[ \begin{array}{cc} SVGA\\K=2 \\C=256\end{array}\right]$ $\times$3             & $\left[ \begin{array}{cc} SVGA\\K=2 \\C=400\end{array}\right]$ $\times$3            & $\left[ \begin{array}{cc} SVGA\\K=2 \\C=464\end{array}\right]$ $\times$5            \\[16pt] \hline 
Head                     & 1 $\times$ 1                 & Pooling \& MLP             & Pooling \& MLP            & Pooling \& MLP            & Pooling \& MLP            \\ \hline
\end{tabular}
\label{table_of_arch}
\end{table*}

Research in computer vision for mobile applications has seen rapid growth, leading to hybrid architectures using CNNs for learning spatially local representations and vision transformers (ViT) for learning global representations \cite{MobileViT}. Current ViG models are not suited for mobile tasks, as they are inefficient and slow when running on mobile devices. The concepts learned from the design of CNN and ViT models can be explored to determine whether CNN-GNN hybrid models can provide the speed of CNN-based models along with the accuracy of ViT-based models. In this work, we investigate hybrid CNN-GNN architectures for computer vision on mobile devices and develop a graph-based attention mechanism that can compete with existing efficient architectures. We summarize our contributions as follows:

\begin{enumerate}
    \item We propose a new graph-based sparse attention method designed for mobile vision applications. We call our attention method Sparse Vision Graph Attention (SVGA). Our method is lightweight as it does not require reshaping and incurs little overhead in graph construction as compared to previous methods.
    \item We propose a novel mobile CNN-GNN architecture for vision tasks using our proposed SVGA, max-relative graph convolution \cite{maxrel}, and concepts from mobile CNN and mobile vision transformer architectures \cite{MobileNet, MobileViT} that we call MobileViG.
    \item Our proposed model, MobileViG, matches or beats existing vision graph neural network (ViG), mobile convolutional neural network (CNN), and mobile vision transformer (ViT) architectures in terms of accuracy and/or speed on three representative vision tasks: ImageNet image classification, COCO object detection, and COCO instance segmentation.
\end{enumerate}

To the best of our knowledge, we are the first to investigate hybrid CNN-GNN architectures for mobile vision applications. Our proposed SVGA attention method and MobileViG architecture open a new path of exploration for state-of-the-art mobile architectures and ViG architectures.

This paper is structured as follows. Section 2 covers related work in the ViG and mobile architecture space. Section 3 describes the design methodology behind SVGA and the MobileViG architecture. Section 4 describes experimental setup and results for ImageNet-1k image classification, COCO object detection, and COCO instance segmentation. Lastly, Section 5 concludes the paper and suggests future work with ViGs in mobile architecture design.

\section{Related Work}
ViG \cite{Vision_GNN} is proposed as an alternative to CNNs and ViTs due to its capacity to represent image data in a more flexible format. ViG represents images through using the KNN algorithm \cite{Vision_GNN}, where each pixel in the image attends to similar pixels. ViG achieves comparable performance to popular ViT models, DeiT \cite{Deit} and SwinTransformer \cite{liu2021swin}, suggesting it is worth further investigations.

Despite the success of ViT-based models in vision tasks, they are still slower when compared to lightweight CNN-based models \cite{EfficientFormer}, in contrast CNN-based models lack the global receptive field of ViT-based models. Thus, ViG-based models may be a possible solution by providing speeds faster than ViT-based models and accuracies higher than CNN-based models. To the best of our knowledge, there are no works on mobile ViGs at this time; however, there are many existing works in the mobile CNN and hybrid model space. We classify mobile architecture designs into two primary categories: convolutional neural network (CNN) models and hybrid CNN-ViT models, which blend elements of CNNs and ViTs.

\begin{figure*}[t]
\centering
\includegraphics[scale=0.713]{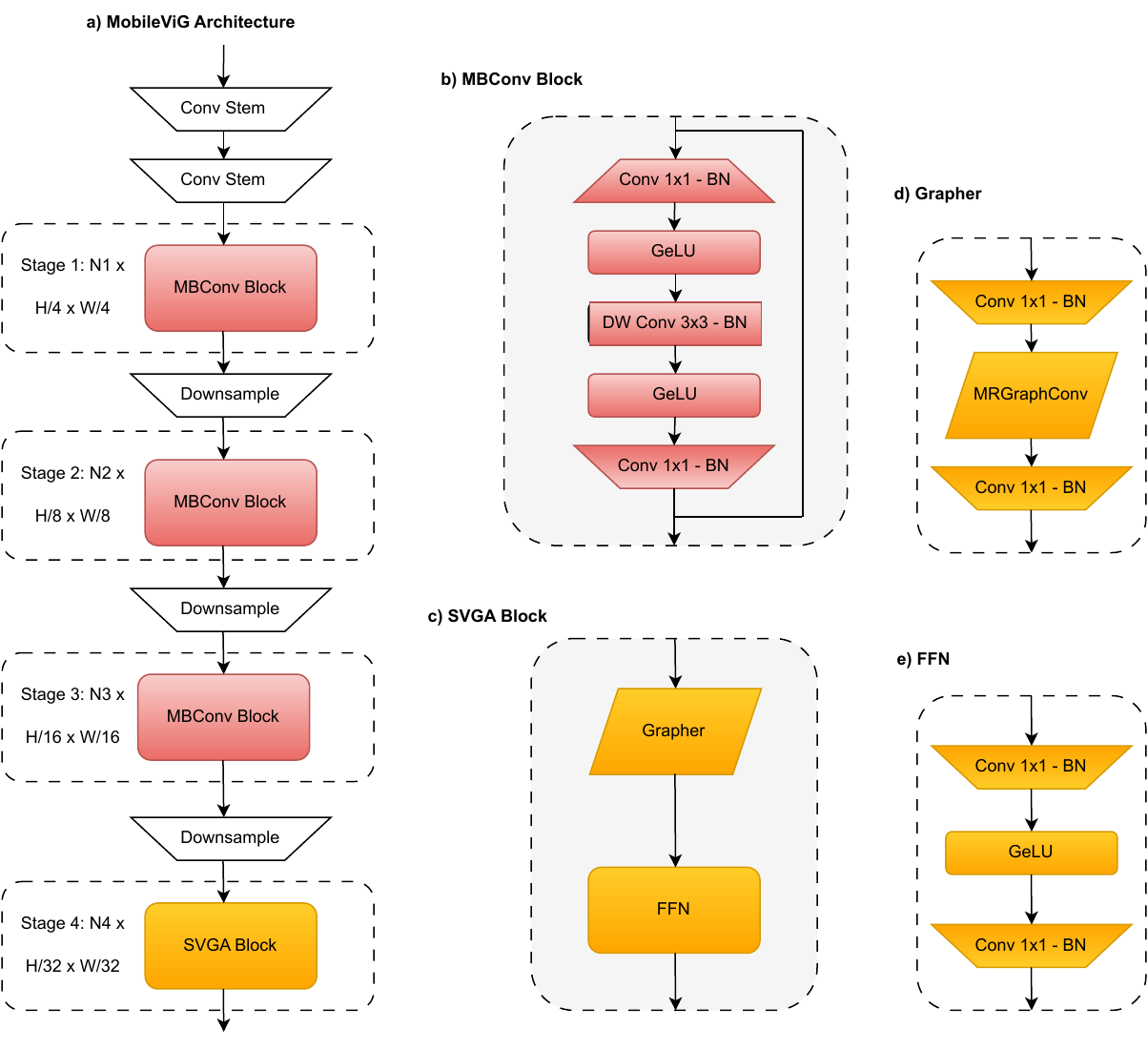}
\caption{MobileViG architecture. (a) network architecture showing the stages and layers, where N1, N2, N3, and N4 represent the number of those blocks in the MobileViG-Ti, S, M, and B configurations (Section 3.3). (b) MBConv Block (Section 3.3). (c) SVGA Block (Section 3.2). (d) \& (e) Grapher and FFN (Section 3.2).}
\label{fig:MobileViG_Architecture}
\end{figure*}

The MobileNetv2 \cite{MobileNetv2} and EfficientNet \cite{tan2019efficientnet, tan2021efficientnetv2} families of CNN-based architectures are some of the first mobile models to see success in common image tasks. These models are lightweight with fast inference speeds. However, purely CNN-based models have steadily been replaced by hybrid competitors.

There are a vast number of hybrid mobile models, including MobileViTv2 \cite{MobileViTv2}, EdgeViT \cite{pan2022edgevits} LeViT \cite{graham2021levit}, and EfficientFormerv2 \cite{EfficientFormerv2}. These hybrid models consistently beat MobileNetv2 in image classification, object detection, and instance segmentation tasks, but some of these models do not always perform as well in terms of latency. The latency difference can be tied to the inclusion of ViT blocks, which have traditionally been slower on mobile hardware. To improve this state of affairs we propose MobileViG, which provides speeds comparable to MobileNetv2\cite{MobileNetv2} and accuracies comparable to EfficientFormer \cite{EfficientFormer}.

\section{Methodology}

In this section, we describe the SVGA algorithm and provide details on the MobileViG architecture design. More precisely, Section 3.1 describes the SVGA algorithm. Section 3.2 explains how we adapt the Grapher module from ViG \cite{Vision_GNN} to create the SVGA block. Section 3.3 describes how we combine the SVGA blocks along with inverted residual blocks for local processing to create MobileViG-Ti, MobileViG-S, MobileViG-M, and MobileViG-B.

\subsection{Sparse Vision Graph Attention}
We propose Sparse Vision Graph Attention (SVGA) as a mobile-friendly alternative to KNN graph attention from Vision GNN \cite{Vision_GNN}. The KNN-based graph attention introduces two non-mobile-friendly components, KNN computation and input reshaping, that we remove with SVGA. 

In greater detail, the KNN computation is required for every input image, since the nearest neighbors of each pixel cannot be known ahead of time. This results in a graph with seemingly random connections as seen in Figure \ref{fig:KNN_ViG}a. Due to the unstructured nature of KNN, the authors of \cite{Vision_GNN} reshape the input image from a 4D to 3D tensor, allowing them to properly align the features of connected pixels for graph convolution. Following the graph convolution, the input must be reshaped from 3D back to 4D for subsequent convolutional layers. Thus, KNN-based attention requires the KNN computation and two reshaping operations, both of which are costly on mobile devices.

To remove the overhead of the KNN computation and reshaping operations, SVGA assumes a fixed graph, where each pixel is connected to every $K^{th}$ pixel in its row and column. For example, given an 8$\times$8 image and $K=2$, the top left pixel would be connected to every second pixel across its row and every second pixel down its column as seen in Figure \ref{fig:KNN_ViG}b. This same pattern is repeated for every pixel in the input image. Since the graph has a fixed structure (i.e., each pixel will have the same connections for all 8$\times$8 input images), the input image does not have to be reshaped to perform the graph convolution.

Instead, it can be implemented using rolling operations across the two image dimensions, denoted as $roll_{right}$ and $roll_{down}$ in Algorithm \ref{alg:cap}. The first parameter to the $roll$ operation is the input to roll, and the second is the distance to roll in the $right$ or $down$ direction. Using the example from Figure \ref{fig:KNN_ViG}b where $K=2$, the top left pixel can be aligned with every second pixel in its row by rolling the image twice to the right, four times to the right, and six times to the right. The same can be done for every second pixel in its column, except by rolling down. Note that since every pixel is connected in the same way, the rolling operations used to align the top left pixel with its connections simultaneously align every other pixel in the image with its connections. In MobileViG, graph convolution is performed using max-relative graph convolution (MRConv). Therefore, after every $roll_{right}$ and $roll_{down}$ operation, the difference between the original input image and the rolled version is computed, denoted as $X_{r}$ and $X_{c}$ in Algorithm \ref{alg:cap}, and the $max$ operation is taken element wise and stored in $X_j$, also denoted in Algorithm \ref{alg:cap}. After completing the rolling and max-relative operations, a final $Conv2d$ is performed. Through this approach, SVGA trades the KNN computation for cheaper rolling operations, consequently not requiring reshaping to perform the graph convolution.

We note that SVGA eschews the representation flexibility of KNN in favor of being mobile friendly.

\begin{algorithm}
\caption{SVGA with MRConv}\label{alg:cap}
\begin{algorithmic}
\Require $K$, the distance between connections; $H,W$, the image resolution; $X$, the input image; $m$, controls the distance of each roll

\State $m \gets 0$
\While{$mK < H$}
    \State $X_c \gets X-roll_{down}(X, mK)$     \Comment{get relative features}
    \State $X_j \gets max(X_c, X_j)$ \Comment{keep max relative features}
    \State $m \gets m+1$            
\EndWhile
\State $m \gets 0$
\While{$mK < W$}
    \State $X_r \gets X-roll_{right}(X, mK)$
    \State $X_j \gets max(X_r, X_j)$
    \State $m \gets m+1$
\EndWhile
\State \Return $Conv2d(Concat(X,X_j))$
\end{algorithmic}
\end{algorithm}

\subsection{SVGA Block}
We insert SVGA and the updated MRConv layer into the Grapher block proposed in Vision GNN \cite{Vision_GNN}. Given an input feature $X\in\mathbb{R}^{N \times N}$, the updated Grapher is expressed as
\begin{equation}
\tag{1}
Y=\sigma(MRConv(XW_{in}))W_{out}+X
\end{equation}
where $Y\in\mathbb{R}^{N \times N}$, $W_{in}$ and $W_{out}$ are fully connected layer weights, and $\sigma$ is a GeLU activation. We also change the number of filter groups from 4 (the value used in Vision GNN \cite{Vision_GNN}) to 1 in the MRConv step to increase the expressive potential of the MRConv layer without a noticeable increase in latency. The updated Grapher module is visually depicted in Figure \ref{fig:MobileViG_Architecture}d

Following the updated Grapher, we use the feed-forward network (FFN) module as proposed in Vision GNN \cite{Vision_GNN} and shown in Figure \ref{fig:MobileViG_Architecture}e The FFN module is a two layer MLP expressed as
\begin{equation}
\tag{2}
Z=\sigma(XW_{1})W_{2}+Y
\end{equation}
where $Z\in\mathbb{R}^{N \times N}$, $W_{1}$ and $W_{2}$ are fully connected layer weights, and $\sigma$ is once again GeLU. We call this combination of updated Grapher and FFN an SVGA block, as shown in Figure \ref{fig:MobileViG_Architecture}c.

\subsection{MobileViG Architecture}

The MobileViG architecture shown in Figure \ref{fig:MobileViG_Architecture}a is composed of a convolutional stem, followed by three stages of inverted residual blocks (MBConv) with an expansion ratio of four for local processing as proposed in MobileNetv2 \cite{MobileNetv2}. Within the MBConv blocks, we swap ReLU6 for GeLU as it has been shown to improve performance in computer vision tasks \cite{ViT, EfficientFormerv2}. The MBConv blocks consist of a 1$\times$1 convolution plus batch normalization (BN) and GeLU, a depth-wise 3$\times$3 convolution plus BN and GeLU, and lastly a 1$\times$1 convolution plus BN and a residual connection as seen in Figure \ref{fig:MobileViG_Architecture}b. Following the MBConv blocks we have one stage of SVGA blocks to capture global information as seen in Figure \ref{fig:MobileViG_Architecture}a. We also have a convolutional head after the SVGA blocks for classification. After each MBConv stage, a downsampling step halves the input resolution and expands the channel dimension. Each stage is composed of multiple MBConv or SVGA blocks, where the number of repetitions is changed depending on model size. The channel dimensions and number of blocks repeated per stage for MobileViG-Ti, MobileViG-S, MobileViG-M, and MobileViG-B can be seen in Table \ref{table_of_arch}.

\begin{table}[tb!]
\def\arraystretch{1.3}
\caption{Top-1 accuracy on ImageNet-1k classification and iPhone13 Mini GPU latency for PyramidViG and MobileViG. Bold entries indicate results obtained using MobileViG and SVGA in this paper.}
\large
\centering
\resizebox{\columnwidth}{!}{
 \begin{tabular}{|c|c|c|c|c|}
 \toprule[\heavyrulewidth]\toprule[\heavyrulewidth]
\textbf{Model} & \textbf{Params (M)} & \textbf{GMACs} & \textbf{GPU Latency (ms)} & \textbf{Top-1 (\%)}  \\ \hlineB{5}

PViG-Ti \cite{Vision_GNN}          & 10.7     & 1.7 & 81.2 & 78.2       \\ \hline
PViG-S \cite{Vision_GNN}           & 27.3     & 4.6 & 111 & 82.1      \\ \hline
PViG-M \cite{Vision_GNN}         & 51.7     & 8.9 & 171 &83.1      \\ \hline
PViG-B \cite{Vision_GNN}            & 92.6 & 16.8     & 242 & 83.7     \\ \hlineB{5}
\textbf{MViG-Ti (Ours)}        & \textbf{5.2}     & \textbf{0.7} &  \textbf{18.0} & \textbf{75.7}       \\ \hline
\textbf{MViG-S (Ours)}       & \textbf{7.2}     & \textbf{1.0} & \textbf{28.7} & \textbf{78.2}      \\ \hline
\textbf{MViG-M (Ours)}        & \textbf{14.0}     & \textbf{1.5} & \textbf{33.2} & \textbf{80.6}      \\ \hline
\textbf{MViG-B (Ours)}       & \textbf{26.7}     & \textbf{2.8} & \textbf{53.4} & \textbf{82.6}      \\ \hlineB{5}

\bottomrule[\heavyrulewidth]
\end{tabular}}
\label{Pyramid_Classification_Results}
\end{table}

\begin{table*}[ht]
\def\arraystretch{1.3}
\caption{Results of MobileViG and other mobile architectures on ImageNet-1k. Latency is reported on the NPU and GPU of the iPhone13 Mini. Models are compiled with CoreML. Bold entries indicate results obtained using MobileViG and SVGA in this paper. }
\centering
\begin{tabular}[t]{|c|c|c|c|c|c|c|c|c|}
\hline
\multirow{2}{*}{\textbf{Model}} & \multirow{2}{*}{\textbf{Type}} & \multirow{2}{*}{\textbf{Params (M)}} & \multirow{2}{*}{\textbf{GMACs}} & \multicolumn{2}{c|}{\textbf{Latency (ms)}} & \multirow{2}{*}{\textbf{Epochs}} & \multirow{2}{*}{\textbf{Top-1 (\%)}}  \\ \cline{5-6}
                                &               &           &     & \textbf{NPU}  & \textbf{GPU}   &     &          \\ \hline
MobileNetV2x1.0 \cite{MobileNetv2}                & CONV          & 3.5       & 0.3   & 0.81  & 13.0        & 300 & 71.8     \\ \hline
MobileViTv2-1.0 \cite{MobileViTv2}                & Hybrid        & 4.9       & 1.8   & 3.13  & 40.2      & 300 & 78.1       \\ \hline
EfficientFormerV2-S0 \cite{EfficientFormerv2}           & Hybrid        & 3.5       & 0.4   & 0.85  & 19.0       & 300 & 75.7     \\ \hline
EdgeViT-XXS \cite{pan2022edgevits}                    & Hybrid        & 4.1       & 0.6   & -     & 25.0       & 300 & 74.4     \\ \hline
\textbf{MobileViG-Ti (Ours)}    & \textbf{CNN-GNN}        & \textbf{5.2}       & \textbf{0.7}   & \textbf{0.78}  & \textbf{18.0}    & \textbf{300} & \textbf{75.7}     \\ \hlineB{5}
MobileNetV2x1.4 \cite{MobileNetv2}                & CONV          & 6.1       & 0.6   & 1.02  & 14.8       & 300 & 74.7     \\ \hline
EfficientNet-B0 \cite{tan2019efficientnet}                & CONV          & 5.3       & 0.4   & 1.89  & 17.9        & 300 & 77.7     \\ \hline
DeiT-T \cite{Deit}                         & Attention     & 5.9       & 1.2   & 8.60  & 27.3       & 300 & 74.5     \\ \hline
EdgeViT-XS \cite{pan2022edgevits}                     & Hybrid        & 6.7       & 1.1   & -     & 36.9       & 300 & 77.5     \\ \hline
EfficientFormerV2-S1 \cite{EfficientFormerv2}           & Hybrid        & 6.1       & 0.7   & 0.93  & 31.7       & 300 & 79.0     \\ \hline
LeViT-128S \cite{graham2021levit}                     & Hybrid        & 7.8       & 0.3   & 7.63  & 8.09       & 1000 & 76.6     \\ \hline
\textbf{MobileViG-S (Ours)}     & \textbf{CNN-GNN}        & \textbf{7.2}       & \textbf{1.0}     & \textbf{0.99}  & \textbf{28.7}     & \textbf{300} & \textbf{78.2}       \\ \hlineB{5}
ResNet18 & CONV & 11.7 & 1.82 & 1.20 & 15.7 & 300 & 69.7 \\ \hline
MobileViTv2-1.5 \cite{MobileViTv2}                & Hybrid        & 10.6      & 4.0   & 4.52  & 70.0       & 300 & 80.4    \\  \hline
EfficientNet-B3 \cite{tan2019efficientnet}                & CONV          & 12.2      & 2.0   & 5.46  & 61.4       & 300 & 82.2     \\ \hline
PoolFormer-s12 \cite{MetaFormer}            & Pool          & 12.0      & 2.0   & 1.47  & 91.7       & 300 & 77.2     \\ \hline
LeViT-192 \cite{graham2021levit}            & Hybrid        & 10.9      & 0.7   & 41.8  & 13.0       & 1000 & 80.0     \\ \hline
EdgeViT-S \cite{pan2022edgevits}                    & Hybrid        & 11.1      & 1.9   & -     & 57.5        & 300 & 81.0     \\ \hline
EfficientFormerV2-S2 \cite{EfficientFormerv2}           & Hybrid        & 12.6      & 1.3   & 1.42  & 60.0      & 300 & 81.6     \\ \hline
EfficientFormer-L1 \cite{EfficientFormer}             & Hybrid        & 12.3      & 1.3   & 1.18  & 18.0        & 300 & 79.2     \\ \hline
\textbf{MobileViG-M (Ours)}     & \textbf{CNN-GNN}        & \textbf{14.0}         & \textbf{1.5}     & \textbf{1.38}  & \textbf{33.2}          & \textbf{300} & \textbf{80.6}       \\ \hlineB{5}
ResNet50 \cite{Resnet}                       & CONV          & 25.6      & 4.1   & 2.29  & 38.2        & 300 & 80.4      \\ \hline
ConvNext-T \cite{liu2022convnet}                     & CONV          & 28.6      & 7.4   & 147   & 227        & 300 & 82.7      \\ \hline
MobileViTv2-2.0 \cite{MobileViTv2}                 & Hybrid        & 18.5      & 7.5   & 6.13  & 128         & 300 & 81.2        \\ \hline
PoolFormer-s24 \cite{MetaFormer}                 & Pool          & 21.0      & 3.6   & 2.48  & 177         & 300 & 80.3      \\ \hline 
PoolFormer-s36 \cite{MetaFormer}                 & Pool          & 31.0      & 5.2   & 3.40  & 266        & 300 & 81.4      \\ \hline
PoolFormer-m36 \cite{MetaFormer}                 & Pool          & 56.0      & 8.8   & 5.73  & 343        & 300 & 82.1      \\ \hline
DeiT-S \cite{Deit}                         & Attention     & 22.5      & 4.5   & 13.7  & 76.9       & 300 & 81.2      \\ \hline
Swin-T \cite{liu2021swin}                      & Attention     & 29.0      & 4.5   & -     & -        & 300 & 81.4      \\ \hline
LeViT-256 \cite{graham2021levit}                       & Hybrid        & 18.9      & 1.1   & 48.5  & 18.7        & 1000 & 81.6     \\ \hline
LeViT-384 \cite{graham2021levit}                      & Hybrid        & 39.1      & 2.4   & 62.0  & 30.7        & 1000 & 82.6     \\ \hline
EfficientFormerV2-L \cite{EfficientFormer}             & Hybrid        & 26.1      & 2.6   & 2.36  & 83.7      & 300 & 83.3      \\ \hline
EfficientFormer-L3 \cite{EfficientFormer}              & Hybrid        & 31.3      & 3.9   & 2.77  & 38.1        & 300 & 82.4      \\ \hline
EfficientFormer-L7 \cite{EfficientFormer}              & Hybrid        & 82.1      & 10.2  & 6.87  & 83.3       & 300 & 83.3     \\ \hline
\textbf{MobileViG-B (Ours)}     & \textbf{CNN-GNN}        & \textbf{26.7}         & \textbf{2.8}     & \textbf{2.30}  & \textbf{53.4}         & \textbf{300} & \textbf{82.6}       \\ \hlineB{5}
\end{tabular}
\label{Mobile_Classification_Results}
\end{table*}

\begin{table*}[ht]
\def\arraystretch{1.2}
\caption{Object detection and instance segmentation results of MobileViG and other backbones on MS COCO 2017. Bold entries indicate results obtained using MobileViG and SVGA in this paper.}
\centering
\begin{tabular}[t]{|c|c|c|c|c|c|c|c|c|c|c|}
\hline
\textbf{Backbone} & \textbf{Params (M)} & \textbf{APb} & \textbf{APb50} & \textbf{APb75} & \textbf{APm} & \textbf{APm50} & \textbf{APm75}  \\ \hline
ResNet18 \cite{Resnet}        & 11.7  & 34.0 & 54.0 & 36.7 & 31.2 & 51.0 & 32.7     \\ \hline
EfficientFormer-L1 \cite{EfficientFormer}   & 12.3 & 37.9 & 60.3 & 41.0 & 35.4 & 57.3 & 37.3        \\ \hline
PoolFormer-S12 \cite{MetaFormer}   & 12.0  & 37.3 & 59.0 & 40.1 & 34.6 & 55.8 & 36.9          \\ \hline
\textbf{MobileViG-M (Ours)}    & \textbf{14.0} & \textbf{41.3} & \textbf{62.8} & \textbf{45.1} & \textbf{38.1} & \textbf{60.1} & \textbf{40.8}    \\ \hlineB{5}
ResNet50 \cite{Resnet}            & 25.5 & 38.0 & 58.6 & 41.4 & 34.4 & 55.1 & 36.7     \\ \hline
EfficientFormer-L3 \cite{EfficientFormer}    & 31.3 & 41.4 & 63.9 & 44.7 & 38.1 & 61.0 & 40.4      \\ \hline
PoolFormer-S24 \cite{MetaFormer}          & 21.0 & 40.1 & 62.2 & 43.4 & 37.0 & 59.1 & 39.6    \\ \hline
PVT-Small \cite{wang2021pyramid}         & 24.5 & 40.4 & 62.9 & 43.8 & 37.8 & 60.1 & 40.3    \\ \hline
\textbf{MobileViG-B (Ours)}    & \textbf{26.7} & \textbf{42.0} & \textbf{64.3} & \textbf{46.0} & \textbf{38.9} & \textbf{61.4} & \textbf{41.6}    \\ \hlineB{5}
\end{tabular}
\label{Object_Detection_Results}
\end{table*}

\section{Experimental Results}

We compare MobileViG to ViG \cite{Vision_GNN} and show its superior performance in terms of latency, model size, and image classification accuracy on ImageNet-1k \cite{imagenet1k} in Table \ref{Pyramid_Classification_Results}. We also compare MobileViG to several mobile models and show that, for each model, it has superior or comparable performance in terms of accuracy and latency in Table \ref{Mobile_Classification_Results}.

\subsection{Image Classification}

We implement the model using PyTorch 1.12 \cite{paszke2019pytorch} and Timm library \cite{timm}.We use 8 NVIDIA A100 GPUs to train each model, with an effective batch size of 1024. The models are trained from scratch for 300 epochs on ImageNet-1K \cite{imagenet1k} with AdamW optimizer  \cite{AdamW}. Learning rate is set to 2e-3 with cosine annealing schedule. We use a standard image resolution, 224 × 224, for both training and testing. Similar to DeiT \cite{Deit}, we perform knowledge distillation using RegNetY-16GF \cite{RegNetY} with 82.9\% top-1 accuracy. For data augmentation we use RandAugment, Mixup, Cutmix, random erasing, and repeated augment.

We use an iPhone 13 Mini (iOS 16) to benchmark latency on NPU and GPU. The models are compiled with CoreML and latency is averaged over 1000 predictions \cite{coreml}.

As seen in Table \ref{Pyramid_Classification_Results}, for a similar number of parameters, MobileViG outperforms Pyramid ViG \cite{Vision_GNN} both in accuracy and GPU latency. For example, for 3.5 M fewer parameters, MobileViG-S matches Pyramid ViG-Ti in top-1 accuracy, while being 2.83$\times$ faster. Additionally, for 0.6 M fewer parameters, MobileViG-B beats Pyramid ViG-S by 0.5\% in top-1 accuracy, while being 2.08$\times$ faster.

When compared to mobile models in Table \ref{Mobile_Classification_Results}, MobileViG consistently beats every model in at least NPU latency, GPU latency, or accuracy. MobileViG-Ti is faster than MobileNetv2 with 3.9\% higher top-1 accuracy. It also matches EfficientFormerv2 \cite{EfficientFormerv2} in top-1 while having a slight edge in NPU and GPU latency. MobileViG-S is nearly 2x faster than EfficientNet-B0 \cite{tan2019efficientnet} in NPU latency and has 0.5\% higher top-1 accuracy. Compared to MobileViTv2-1.5 \cite{MobileViTv2}, MobileViG-M is over 3x faster in NPU latency and 2x faster in GPU latency with 0.2\% higher top-1 accuracy. Additionally, MobileViG-B is 6x faster than DeiT-S and is able to beat both DeiT-S and Swin-Tiny in top-1 accuracy.

\subsection{Object Detection and Instance Segmentation}

We evaluate MobileViG on object detection and instance segmentation tasks to further prove the potential of SVGA. We integrate MobileViG as a backbone in the Mask-RCNN framework \cite{mask_r_cnn} and experiment using the MS COCO 2017 dataset \cite{coco}. We implement the backbone using PyTorch 1.12 \cite{paszke2019pytorch} and Timm library \cite{timm}, and use 4 NVIDIA RTX A6000 GPUs to train our models. We initialize the model with pretrained ImageNet-1k weights from 300 epochs of training, use AdamW \cite{AdamW} optimizer with an initial learning rate of 2e-4 and train the model for 12 epochs with a standard resolution (1333 X 800) following the process of Next-ViT, EfficientFormer, and EfficientFormerV2 \cite{li2022next, EfficientFormer, EfficientFormerv2}.

As seen in Table \ref{Object_Detection_Results}, with similar model size MobileViG outperforms ResNet, PoolFormer, EfficientFormer, and PVT in terms of either parameters or improved average precision (AP) on object detection and/or instance segmentation. The medium size MobileViG-M model gets 41.3 APbox, 62.8 APbox when 50 Intersection over Union (IoU), and 45.1 APbox when 75 IoU on the object detection task. MobileViG-M gets 38.1 APmask, 60.1 APmask when 50 IoU, and 40.8 APmask when 75 IoU for the instance segmentation task. The big size MobileViG-B model gets 42.0 APbox, 64.3 APbox when 50 IoU, and 46.0 APbox when 75 IoU on the object detection task. MobileViG-B gets 38.9 APmask, 61.4 APmask when 50 IoU, and 41.6 APmask when 75 IoU on the instance segmentation task. The strong performance of MobileViG on object detection and instance segmentation shows that MobileViG generalizes well as a backbone for different tasks in computer vision.

The design of MobileViG is partly inspired by the designs of Pyramid ViG \cite{Vision_GNN}, EfficientFormer \cite{EfficientFormer}, and the MetaFormer concept \cite{MetaFormer}. The results achieved in MobileViG demonstrate that hybrid CNN-GNN architectures are a viable alternative to CNN, ViT, and hybrid CNN-ViT designs. Hybrid CNN-GNN architectures can provide the speed of CNN-based models along with the accuracy of ViT models making them an ideal candidate for high accuracy mobile architecture designs. Further explorations of hybrid CNN-GNN architectures for mobile computer vision tasks can improve on the MobileViG concept and introduce new state-of-the-art architectures.

\section{Conclusion}

In this work, we have proposed a graph-based attention mechanism, Sparse Vision Graph Attention (SVGA), and MobileViG, a competitive mobile vision architecture that uses SVGA. SVGA does not require reshaping and allows for the graph structure to be known prior to inference, unlike previous methods. We use inverted residual blocks, max-relative graph convolution, and feed-forward network layers to create MobileViG, a hybrid CNN-GNN architecture, that achieves competitive results on image classification, object detection, and instance segmentation tasks. MobileViG outperforms existing ViG models and many existing mobile models, including MobileNetv2, in terms of accuracy and latency. Future research on mobile architectures can further explore the potential of GNN-based models on resource-constrained devices for IoT applications.

\noindent 

{\small
\bibliographystyle{ieee_fullname}
\bibliography{egbib}

\begin{thebibliography}{10}\itemsep=-1pt

\bibitem{coreml}
{Apple Inc.}
\newblock Core ml.
\newblock \url{https://developer.apple.com/documentation/coreml}, 2017.

\bibitem{cyclemlp}
Shoufa Chen, Enze Xie, Chongjian Ge, Ding Liang, and Ping Luo.
\newblock Cyclemlp: A mlp-like architecture for dense prediction.
\newblock {\em arXiv preprint arXiv:2107.10224}, 2021.

\bibitem{imagenet1k}
Jia Deng, Wei Dong, Richard Socher, Li-Jia Li, Kai Li, and Li Fei-Fei.
\newblock Imagenet: A large-scale hierarchical image database.
\newblock In {\em 2009 IEEE Conference on Computer Vision and Pattern
  Recognition}, pages 248--255, 2009.

\bibitem{ViT}
Alexey Dosovitskiy et~al.
\newblock An image is worth 16x16 words: Transformers for image recognition at
  scale.
\newblock {\em arXiv preprint arXiv:2010.11929}, 2020.

\bibitem{gaudelet2021utilizing}
Thomas Gaudelet, Ben Day, Arian~R Jamasb, Jyothish Soman, Cristian Regep,
  Gertrude Liu, Jeremy~BR Hayter, Richard Vickers, Charles Roberts, Jian Tang,
  et~al.
\newblock Utilizing graph machine learning within drug discovery and
  development.
\newblock {\em Briefings in bioinformatics}, 22(6):bbab159, 2021.

\bibitem{graham2021levit}
Benjamin Graham, Alaaeldin El-Nouby, Hugo Touvron, Pierre Stock, Armand Joulin,
  Herv{\'e} J{\'e}gou, and Matthijs Douze.
\newblock Levit: a vision transformer in convnet's clothing for faster
  inference.
\newblock In {\em Proceedings of the IEEE/CVF international conference on
  computer vision}, pages 12259--12269, 2021.

\bibitem{hamilton2017inductive}
Will Hamilton, Zhitao Ying, and Jure Leskovec.
\newblock Inductive representation learning on large graphs.
\newblock {\em Advances in neural information processing systems}, 30, 2017.

\bibitem{Vision_GNN}
Kai Han, Yunhe Wang, Jianyuan Guo, Yehui Tang, and Enhua Wu.
\newblock Vision gnn: An image is worth graph of nodes.
\newblock {\em arXiv preprint arXiv:2206.00272}, 2022.

\bibitem{mask_r_cnn}
Kaiming He, Georgia Gkioxari, Piotr Doll{\'a}r, and Ross Girshick.
\newblock Mask r-cnn.
\newblock In {\em Proceedings of the IEEE international conference on computer
  vision}, pages 2961--2969, 2017.

\bibitem{Resnet}
Kaiming He, Xiangyu Zhang, Shaoqing Ren, and Jian Sun.
\newblock Deep residual learning for image recognition.
\newblock In {\em Proceedings of the IEEE conference on computer vision and
  pattern recognition}, pages 770--778, 2016.

\bibitem{LSTM}
Sepp Hochreiter and J{\"u}rgen Schmidhuber.
\newblock Long short-term memory.
\newblock {\em Neural computation}, 9(8):1735--1780, 1997.

\bibitem{MobileNet}
Andrew~G Howard, Menglong Zhu, Bo Chen, Dmitry Kalenichenko, Weijun Wang,
  Tobias Weyand, Marco Andreetto, and Hartwig Adam.
\newblock Mobilenets: Efficient convolutional neural networks for mobile vision
  applications.
\newblock {\em arXiv preprint arXiv:1704.04861}, 2017.

\bibitem{Densenet}
Gao Huang, Zhuang Liu, Laurens Van Der~Maaten, and Kilian~Q Weinberger.
\newblock Densely connected convolutional networks.
\newblock In {\em Proceedings of the IEEE conference on computer vision and
  pattern recognition}, pages 4700--4708, 2017.

\bibitem{kipf2016semi}
Thomas~N Kipf and Max Welling.
\newblock Semi-supervised classification with graph convolutional networks.
\newblock {\em arXiv preprint arXiv:1609.02907}, 2016.

\bibitem{Alexnet2012}
Alex Krizhevsky, Ilya Sutskever, and Geoffrey~E Hinton.
\newblock Imagenet classification with deep convolutional neural networks.
\newblock {\em NeurIPS}, 2012.

\bibitem{lecun1989backpropagation}
Yann LeCun, Bernhard Boser, John~S Denker, Donnie Henderson, Richard~E Howard,
  Wayne Hubbard, and Lawrence~D Jackel.
\newblock Backpropagation applied to handwritten zip code recognition.
\newblock {\em Neural computation}, 1(4):541--551, 1989.

\bibitem{lecun1998gradient}
Yann LeCun, L{\'e}on Bottou, Yoshua Bengio, and Patrick Haffner.
\newblock Gradient-based learning applied to document recognition.
\newblock {\em Proceedings of the IEEE}, 86(11):2278--2324, 1998.

\bibitem{maxrel}
Guohao Li, Matthias Muller, Ali Thabet, and Bernard Ghanem.
\newblock Deepgcns: Can gcns go as deep as cnns?
\newblock In {\em Proceedings of the IEEE/CVF international conference on
  computer vision}, pages 9267--9276, 2019.

\bibitem{li2022next}
Jiashi Li, Xin Xia, Wei Li, Huixia Li, Xing Wang, Xuefeng Xiao, Rui Wang, Min
  Zheng, and Xin Pan.
\newblock Next-vit: Next generation vision transformer for efficient deployment
  in realistic industrial scenarios.
\newblock {\em arXiv preprint arXiv:2207.05501}, 2022.

\bibitem{EfficientFormerv2}
Yanyu Li, Ju Hu, Yang Wen, Georgios Evangelidis, Kamyar Salahi, Yanzhi Wang,
  Sergey Tulyakov, and Jian Ren.
\newblock Rethinking vision transformers for mobilenet size and speed.
\newblock {\em arXiv preprint arXiv:2212.08059}, 2022.

\bibitem{EfficientFormer}
Yanyu Li, Geng Yuan, Yang Wen, Eric Hu, Georgios Evangelidis, Sergey Tulyakov,
  Yanzhi Wang, and Jian Ren.
\newblock Efficientformer: Vision transformers at mobilenet speed.
\newblock {\em arXiv preprint arXiv:2206.01191}, 2022.

\bibitem{coco}
Tsung-Yi Lin, Michael Maire, Serge Belongie, James Hays, Pietro Perona, Deva
  Ramanan, Piotr Doll{\'a}r, and C~Lawrence Zitnick.
\newblock Microsoft coco: Common objects in context.
\newblock In {\em Computer Vision--ECCV 2014: 13th European Conference, Zurich,
  Switzerland, September 6-12, 2014, Proceedings, Part V 13}, pages 740--755.
  Springer, 2014.

\bibitem{liu2021pick}
Yang Liu, Xiang Ao, Zidi Qin, Jianfeng Chi, Jinghua Feng, Hao Yang, and Qing
  He.
\newblock Pick and choose: a gnn-based imbalanced learning approach for fraud
  detection.
\newblock In {\em Proceedings of the Web Conference 2021}, pages 3168--3177,
  2021.

\bibitem{liu2021swin}
Ze Liu, Yutong Lin, Yue Cao, Han Hu, Yixuan Wei, Zheng Zhang, Stephen Lin, and
  Baining Guo.
\newblock Swin transformer: Hierarchical vision transformer using shifted
  windows.
\newblock In {\em Proceedings of the IEEE/CVF international conference on
  computer vision}, pages 10012--10022, 2021.

\bibitem{liu2022convnet}
Zhuang Liu, Hanzi Mao, Chao-Yuan Wu, Christoph Feichtenhofer, Trevor Darrell,
  and Saining Xie.
\newblock A convnet for the 2020s.
\newblock In {\em Proceedings of the IEEE/CVF Conference on Computer Vision and
  Pattern Recognition}, pages 11976--11986, 2022.

\bibitem{AdamW}
Ilya Loshchilov and Frank Hutter.
\newblock Decoupled weight decay regularization.
\newblock {\em arXiv preprint arXiv:1711.05101}, 2017.

\bibitem{MobileViT}
Sachin Mehta and Mohammad Rastegari.
\newblock Mobilevit: light-weight, general-purpose, and mobile-friendly vision
  transformer.
\newblock {\em arXiv preprint arXiv:2110.02178}, 2021.

\bibitem{MobileViTv2}
Sachin Mehta and Mohammad Rastegari.
\newblock Separable self-attention for mobile vision transformers.
\newblock {\em arXiv preprint arXiv:2206.02680}, 2022.

\bibitem{pan2022edgevits}
Junting Pan, Adrian Bulat, Fuwen Tan, Xiatian Zhu, Lukasz Dudziak, Hongsheng
  Li, Georgios Tzimiropoulos, and Brais Martinez.
\newblock Edgevits: Competing light-weight cnns on mobile devices with vision
  transformers.
\newblock In {\em Computer Vision--ECCV 2022: 17th European Conference, Tel
  Aviv, Israel, October 23--27, 2022, Proceedings, Part XI}, pages 294--311.
  Springer, 2022.

\bibitem{paszke2019pytorch}
Adam Paszke et~al.
\newblock Pytorch: An imperative style, high-performance deep learning library.
\newblock {\em Advances in neural information processing systems}, 32, 2019.

\bibitem{RegNetY}
Ilija Radosavovic, Raj~Prateek Kosaraju, Ross Girshick, Kaiming He, and Piotr
  Doll{\'a}r.
\newblock Designing network design spaces.
\newblock In {\em Proceedings of the IEEE/CVF conference on computer vision and
  pattern recognition}, pages 10428--10436, 2020.

\bibitem{Imagenet2015}
Olga Russakovsky et~al.
\newblock Imagenet large scale visual recognition challenge.
\newblock {\em International journal of computer vision}, 115:211--252, 2015.

\bibitem{MobileNetv2}
Mark Sandler, Andrew Howard, Menglong Zhu, Andrey Zhmoginov, and Liang-Chieh
  Chen.
\newblock Mobilenetv2: Inverted residuals and linear bottlenecks.
\newblock 2018.

\bibitem{VGGNet}
Karen Simonyan and Andrew Zisserman.
\newblock Very deep convolutional networks for large-scale image recognition.
\newblock {\em arXiv preprint arXiv:1409.1556}, 2014.

\bibitem{tan2019efficientnet}
Mingxing Tan and Quoc Le.
\newblock Efficientnet: Rethinking model scaling for convolutional neural
  networks.
\newblock In {\em International conference on machine learning}, pages
  6105--6114. PMLR, 2019.

\bibitem{tan2021efficientnetv2}
Mingxing Tan and Quoc Le.
\newblock Efficientnetv2: Smaller models and faster training.
\newblock In {\em International conference on machine learning}, pages
  10096--10106. PMLR, 2021.

\bibitem{mlpmixer}
Ilya~O Tolstikhin, Neil Houlsby, Alexander Kolesnikov, Lucas Beyer, Xiaohua
  Zhai, Thomas Unterthiner, Jessica Yung, Andreas Steiner, Daniel Keysers,
  Jakob Uszkoreit, et~al.
\newblock Mlp-mixer: An all-mlp architecture for vision.
\newblock {\em Advances in neural information processing systems},
  34:24261--24272, 2021.

\bibitem{resmlp}
Hugo Touvron, Piotr Bojanowski, Mathilde Caron, Matthieu Cord, Alaaeldin
  El-Nouby, Edouard Grave, Gautier Izacard, Armand Joulin, Gabriel Synnaeve,
  Jakob Verbeek, et~al.
\newblock Resmlp: Feedforward networks for image classification with
  data-efficient training.
\newblock {\em IEEE Transactions on Pattern Analysis and Machine Intelligence},
  2022.

\bibitem{Deit}
Hugo Touvron, Matthieu Cord, Matthijs Douze, Francisco Massa, Alexandre
  Sablayrolles, and Herv{\'e} J{\'e}gou.
\newblock Training data-efficient image transformers \& distillation through
  attention.
\newblock In {\em International conference on machine learning}, pages
  10347--10357. PMLR, 2021.

\bibitem{vaswani2017attention}
Ashish Vaswani, Noam Shazeer, Niki Parmar, Jakob Uszkoreit, Llion Jones,
  Aidan~N Gomez, {\L}ukasz Kaiser, and Illia Polosukhin.
\newblock Attention is all you need.
\newblock {\em Advances in neural information processing systems}, 30, 2017.

\bibitem{wang2021pyramid}
Wenhai Wang, Enze Xie, Xiang Li, Deng-Ping Fan, Kaitao Song, Ding Liang, Tong
  Lu, Ping Luo, and Ling Shao.
\newblock Pyramid vision transformer: A versatile backbone for dense prediction
  without convolutions.
\newblock In {\em Proceedings of the IEEE/CVF international conference on
  computer vision}, pages 568--578, 2021.

\bibitem{timm}
Ross Wightman.
\newblock {PyTorch Image Models}.
\newblock \url{https://github.com/rwightman/pytorch-image-models}, 2019.

\bibitem{wu2020graph}
Yongji Wu, Defu Lian, Yiheng Xu, Le Wu, and Enhong Chen.
\newblock Graph convolutional networks with markov random field reasoning for
  social spammer detection.
\newblock In {\em Proceedings of the AAAI conference on artificial
  intelligence}, volume~34, pages 1054--1061, 2020.

\bibitem{MetaFormer}
Weihao Yu, Mi Luo, Pan Zhou, Chenyang Si, Yichen Zhou, Xinchao Wang, Jiashi
  Feng, and Shuicheng Yan.
\newblock Metaformer is actually what you need for vision.
\newblock In {\em Proceedings of the IEEE/CVF Conference on Computer Vision and
  Pattern Recognition}, pages 10819--10829, 2022.

\bibitem{zhou2020graph}
Jie Zhou, Ganqu Cui, Shengding Hu, Zhengyan Zhang, Cheng Yang, Zhiyuan Liu,
  Lifeng Wang, Changcheng Li, and Maosong Sun.
\newblock Graph neural networks: A review of methods and applications.
\newblock {\em AI open}, 1:57--81, 2020.

\end{thebibliography}
}

\end{document}